\documentclass[10pt]{article}

\usepackage[
  letterpaper,
  left=25mm,
  right=25mm,
  top=65.2bp,
  bottom=69bp,
  footskip=30bp
]{geometry}

\usepackage[T1]{fontenc}
\usepackage[utf8]{inputenc}
\usepackage{microtype}

\newcommand{\headfont}{\fontfamily{ptm}\fontseries{b}\selectfont}
\newcommand{\headsize}[2]{\fontsize{#1}{#2}\selectfont}

\usepackage{amsmath,amssymb,amsfonts}
\usepackage{graphicx}
\usepackage{subcaption}
\usepackage{booktabs}
\usepackage{float}
\usepackage[table]{xcolor}
\usepackage[numbers,square,sort&compress]{natbib}
\usepackage{url}
\usepackage[most]{tcolorbox}
\usepackage{titlesec}
\usepackage[font=small,labelsep=space]{caption}
\usepackage[
  colorlinks=true,
  linkcolor=LinkBlue,
  citecolor=LinkBlue,
  urlcolor=LinkBlue
]{hyperref}

\DeclareCaptionFont{head}{\headfont\headsize{7.9bp}{11bp}}
\definecolor{FrontBox}{HTML}{F1F4F7}
\definecolor{LinkBlue}{HTML}{0064E0}
\definecolor{AbstractText}{HTML}{1C2B33}

\titleformat{\section}{\headfont\headsize{12.6bp}{15bp}}{\thesection}{11bp}{}
\titleformat{\subsection}{\headfont\headsize{10.5bp}{13bp}}{\thesubsection}{11bp}{}
\titleformat{\subsubsection}{\headfont\headsize{8.8bp}{12bp}}{\thesubsubsection}{11bp}{}
\titlespacing*{\section}{0pt}{6.6pt plus 2pt}{4.8pt}
\titlespacing*{\subsection}{0pt}{5.7pt plus 2pt}{3pt}
\titlespacing*{\subsubsection}{0pt}{8pt plus 2pt}{2pt}

\newtcolorbox{frontmatterbox}{
  enhanced,
  width=\textwidth,
  colback=FrontBox,
  colframe=FrontBox,
  boxrule=0pt,
  arc=10bp,
  left=17bp,
  right=17bp,
  top=18.1bp,
  bottom=17bp,
  boxsep=0pt,
  before skip=0pt,
  after skip=20.3pt
}

\newcommand{\papertitle}[1]{%
  {\raggedright\headfont\headsize{15bp}{22bp} #1\par}\vspace{10.2pt}}
\newcommand{\paperauthors}[1]{%
  {\raggedright\headfont\headsize{10.8bp}{12bp} #1\par}\vspace{7.5pt}}

\newcommand{\paperabstract}[1]{%
  {\normalfont\normalsize\color{AbstractText} #1\par}\vspace{8.2pt}}

\usepackage{times}
\usepackage{latexsym}
\usepackage{array}
\usepackage{latexsym}
\usepackage{multirow}
\usepackage{booktabs}
\usepackage{makecell} 
\usepackage[table]{xcolor}
\newcommand{\nbatablefont}{\fontencoding{T1}\fontfamily{ptm}\selectfont}

\usepackage{booktabs,tabularx,array,amssymb}
\usepackage[table]{xcolor}
\usepackage{caption}

\definecolor{BenchHeader}{RGB}{244,245,247}
\definecolor{BenchHighlight}{RGB}{235,242,250}

\usepackage{tikz}
\usepackage{graphicx}
\usepackage{placeins}
\definecolor{nbaElo}{RGB}{234,241,249}
\definecolor{nbaWins}{RGB}{237,234,245}
\definecolor{nbaLosses}{RGB}{249,238,227}
\definecolor{nbaCash}{RGB}{233,243,235}
\definecolor{nbaEloBest}{RGB}{210,228,247}
\definecolor{nbaWinsBest}{RGB}{224,217,242}
\definecolor{nbaLossesBest}{RGB}{244,223,200}
\definecolor{nbaCashBest}{RGB}{216,234,218}
\definecolor{nbaCashNegative}{RGB}{238,211,224}
\newcommand{\nbahl}[2]{\begingroup
  \setlength{\fboxsep}{0pt}%
  \colorbox{#1}{\hspace{0.12em}\rule[-0.18em]{0pt}{0.98em}#2\hspace{0.12em}}%
\endgroup}
\DeclareRobustCommand{\goldmedal}{%
  \tikz[x=0.48pt,y=0.48pt,baseline=1.1pt]{%
    \shade[top color={rgb,255:red,74;green,144;blue,249},
           bottom color={rgb,255:red,47;green,99;blue,215}]
      (8,10.5)--(1.1,17.2)
      .. controls (-0.2,18.5) and (0.8,20) .. (2.4,20)
      --(6.3,20).. controls (7.5,20) and (8,19.3) .. (8.6,18.3)
      --(11.9,12.3)--cycle;
    \shade[top color={rgb,255:red,81;green,151;blue,253},
           bottom color={rgb,255:red,48;green,108;blue,225}]
      (8,10.5)--(14.9,17.2)
      .. controls (16.2,18.5) and (15.2,20) .. (13.6,20)
      --(9.7,20).. controls (8.5,20) and (8,19.3) .. (7.4,18.3)
      --(4.1,12.3)--cycle;
    \shade[left color={rgb,255:red,255;green,209;blue,79},
           right color={rgb,255:red,241;green,148;blue,36}]
      (8,7.2) circle[radius=7.1];
    \shade[top color={rgb,255:red,255;green,221;blue,93},
           bottom color={rgb,255:red,249;green,169;blue,45}]
      (7.65,7.5) circle[radius=6.4];
    \shade[left color={rgb,255:red,255;green,185;blue,45},
           right color={rgb,255:red,255;green,207;blue,91}]
      (7.85,7.35) circle[radius=5.35];
    \draw[draw={rgb,255:red,237;green,159;blue,38},line width=0.22pt]
      (7.85,7.35) circle[radius=5.35];
    \draw[draw={rgb,255:red,255;green,235;blue,145},line width=0.28pt,
          line cap=round] (3.6,10.7) arc[start angle=141,end angle=104,radius=5.4];
    \fill[fill={rgb,255:red,169;green,81;blue,30}]
      (7.25,10.2).. controls (7.7,10.5) and (8.25,10.5) .. (8.8,10.3)
      --(8.8,4.15).. controls (8.4,3.9) and (7.7,3.9) .. (7.3,4.15)
      --(7.3,8.6)--(6.55,8.6)--(6.55,9.7)--cycle;
  }%
}

\usepackage{tabularx}  
\usepackage{tcolorbox}
\tcbuselibrary{listings,breakable}
\newcolumntype{C}[1]{>{\centering\arraybackslash}p{#1}}
\usepackage[T1]{fontenc}

\usepackage[utf8]{inputenc}

\usepackage{microtype}

\usepackage{inconsolata}

\usepackage{graphicx}

\usepackage{booktabs}
\usepackage{tabularx}
\usepackage{array}
\usepackage{amssymb}

\usepackage{booktabs}
\usepackage{listings}
\usepackage{caption}
\begin{document}

\begin{frontmatterbox}

\papertitle{GM League Arena: An Interactive Long-Horizon Task Benchmark Based on Professional Sports League Management}

\paperauthors{%
  Junkai Zhou, 
  Ziming Pan, 
  Zhaoyi Zhang
}


\paperabstract{
Long-horizon decision-making under sustained competition requires large language models (LLMs) to balance immediate investment with future resource needs. 
However, local success in multi-step interactions can not adequately capture cumulative decision effects or demonstrate long-term goal achievement under resource constraints and sustained competition.
To address this gap, we introduce \textbf{GM Bench}, a benchmark for managing professional sports teams across seasons under resource constraints and competition among teams.
We build the environment simulators for four sports and add financial constraints, player demands, and market mechanisms to establish a unified task across sports. Diverse rosters of comparable strength are constructed from real-world player data, and each LLM is evaluated as a team general manager to assess its long-term management capabilities. We evaluate competitive performance, cash reserves, and cross-season improvement across four sports. Over five NBA seasons, Claude Opus 5.5 ranks first among 29 LLMs, while GPT-6 Astra and GPT-6.1 Sol combine strong performance with larger cash reserves.
}


\end{frontmatterbox}

\section{Introduction}


In complex environments, large language models need to understand rules, take actions, and revise decisions using feedback. Recent benchmarks span scientific tasks~\citep{scienceagentbench,discoverybench}, web interaction~\citep{browsergym,webarena_verified}, and long-horizon games~\citep{balrog,fmbench}. In long-horizon competition, participant interactions shape resource values and action opportunities, and strategies and demands evolve with the environment; models must balance current investment, future resources, and subsequent choices. Evaluating this ability requires evolving environments that link interactions and resource dynamics to cumulative decision outcomes. 

Sports management provides a concrete setting for studying this process, making it possible to track how the model coordinates current inputs with future returns through a series of decisions: trades and costly signings can strengthen a team in the current season while depleting future assets and limiting roster flexibility. These effects unfold as seasons progress, while rivals continue to improve their rosters, prompting further resource reallocation. We therefore formulate team management across seasons as a long-horizon evaluation task, tracing how interactive decisions shape competitive performance and resources over time to test whether models can turn local interactions into sustainable long-term management outcomes.

We introduce \textbf{GM Bench}, which models team management across seasons with two components: interactive subjects and the simulation environment. The interactive subjects comprise LLM general managers (GMs), players and agents; only LLMs serving as GMs are evaluated. GMs draft, sign, and trade players using visible information while competing for market resources. Players and agents share a general LLM with a fixed configuration to express demands and negotiate. The environment uses the National Basketball Association (NBA) as its primary setting and comprises three modules: \emph{League Rules}, \emph{Player \& Team Analysis}, and \emph{Competition Simulation}. A built-in engine simulates games and enforces rules, while financial constraints and market signals shape management decisions.

\begin{figure*}
    \centering 
    \includegraphics[width=15.5cm]{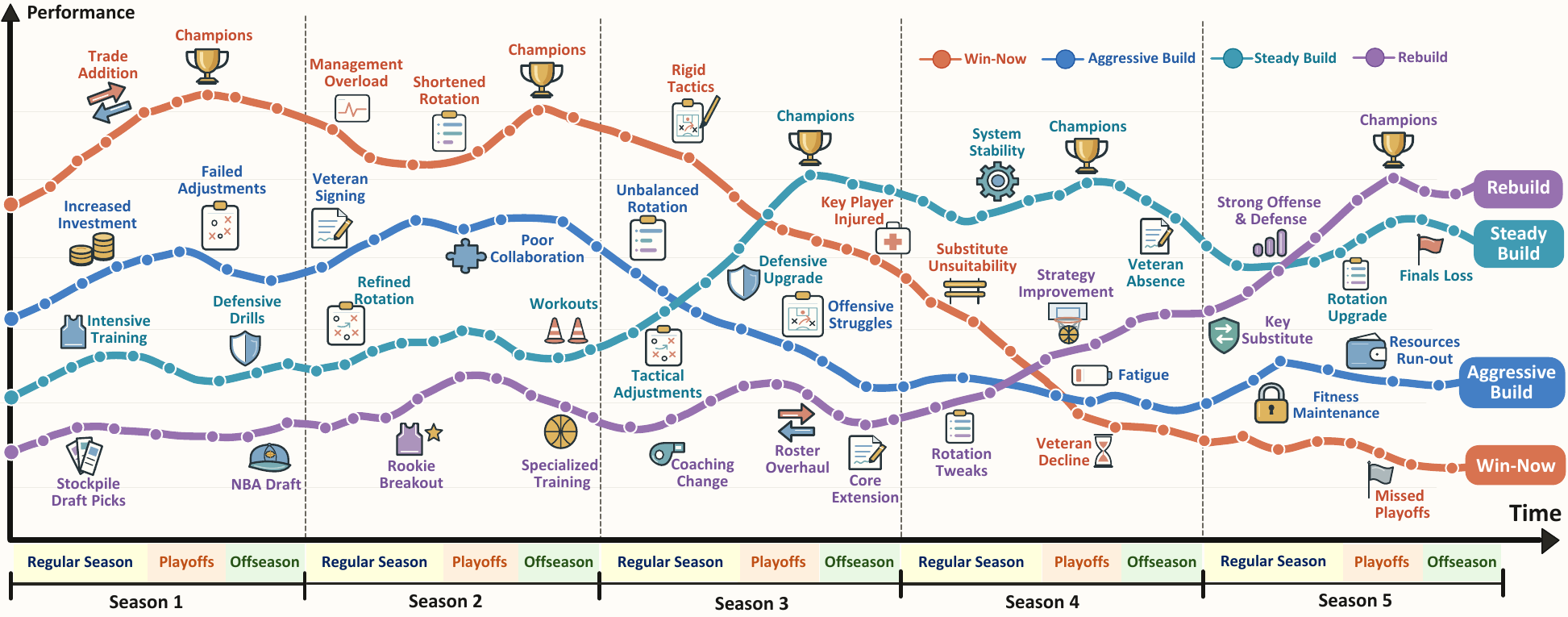}
    \caption{
    Illustrative performance trajectories of four team management strategies over five seasons show that early competitive gains can diminish, whereas rebuilding and sustained investment can yield delayed benefits.}
    \label{fig:fig1}
\end{figure*}

We extend a unified task linking GMs, players, institutional rules, and information channels to hockey, baseball, and American football to assess long-term management under different rules and resources. Real-world player ratings are used to construct five roster types: balanced, single-star, dual-star, offensive, and defensive. Evaluation uses uniform information conditions and comparable initial strength while preserving differences in star configurations and offensive or defensive emphasis. The proposed metrics report competitive performance across seasons, terminal cash, and seasonal trajectories, while action logs are analyzed to assess execution reliability and resource costs.

Experiments across four sports reveal substantial differences in the long-term management capabilities of LLMs. The six leading NBA models remain among the top seven in each additional sport. In the NBA evaluation, for example, Claude Opus 5.5 ranks first and wins three championships over five seasons; GPT-6 Astra and GPT-6.1 Sol combine sustained gains with larger cash reserves. Grok 4.7 achieves stronger competitive returns with less remaining cash, whereas Kimi K3 combines steady improvement with a larger financial buffer. MiMo-V2.6-Pro retains substantial cash despite declining performance. Sonnet faces unaffordable renewals after an NBA championship run. These cases highlight the need to coordinate acquisition, retention, and financing for sustained long-term performance. Our contributions are threefold:
\begin{itemize}
\item We introduce GM Bench across four sports, combining multiple role competition and interactions to evaluate long-horizon resource allocation by LLMs.
\item We develop a protocol jointly reporting competitive performance, terminal resources, and seasonal trajectories, with five roster types of comparable strength from real-world player data for strategy comparisons.
\item We evaluate cross-season management and cross-sport generalization through action logs, and trading diagnostics, analyzing feedback use and long-term decision effects.
\end{itemize}

\section{Task Definition}
\label{sec:formal}

\subsection{Roles and Management Objectives}
Each evaluated LLM serves as a team GM, managing players and assets through selection, trades, and contract renewals. Players share a general LLM for interactions, while the environment handles match simulation and player rotation. The objective is to improve competitive performance across seasons by balancing current roster investment with future resource needs under resource constraints. We model the task as a partially observable stochastic game:
\begin{equation}
\mathcal G=(\mathcal I,\mathcal S,\mathcal A,P,O,\mathbf U,H),
\end{equation}
where $\mathcal I$ is the team set, $\mathcal S$ comprises roster, contract, financial, and market states, and $\mathcal A$ is the joint GM action space. $P$ governs state transitions, $O$ determines information visibility, $\mathbf U$ represents management outcomes, and $H$ is the decision horizon across seasons.

\begin{figure*}
    \centering 
    \includegraphics[width=16cm]{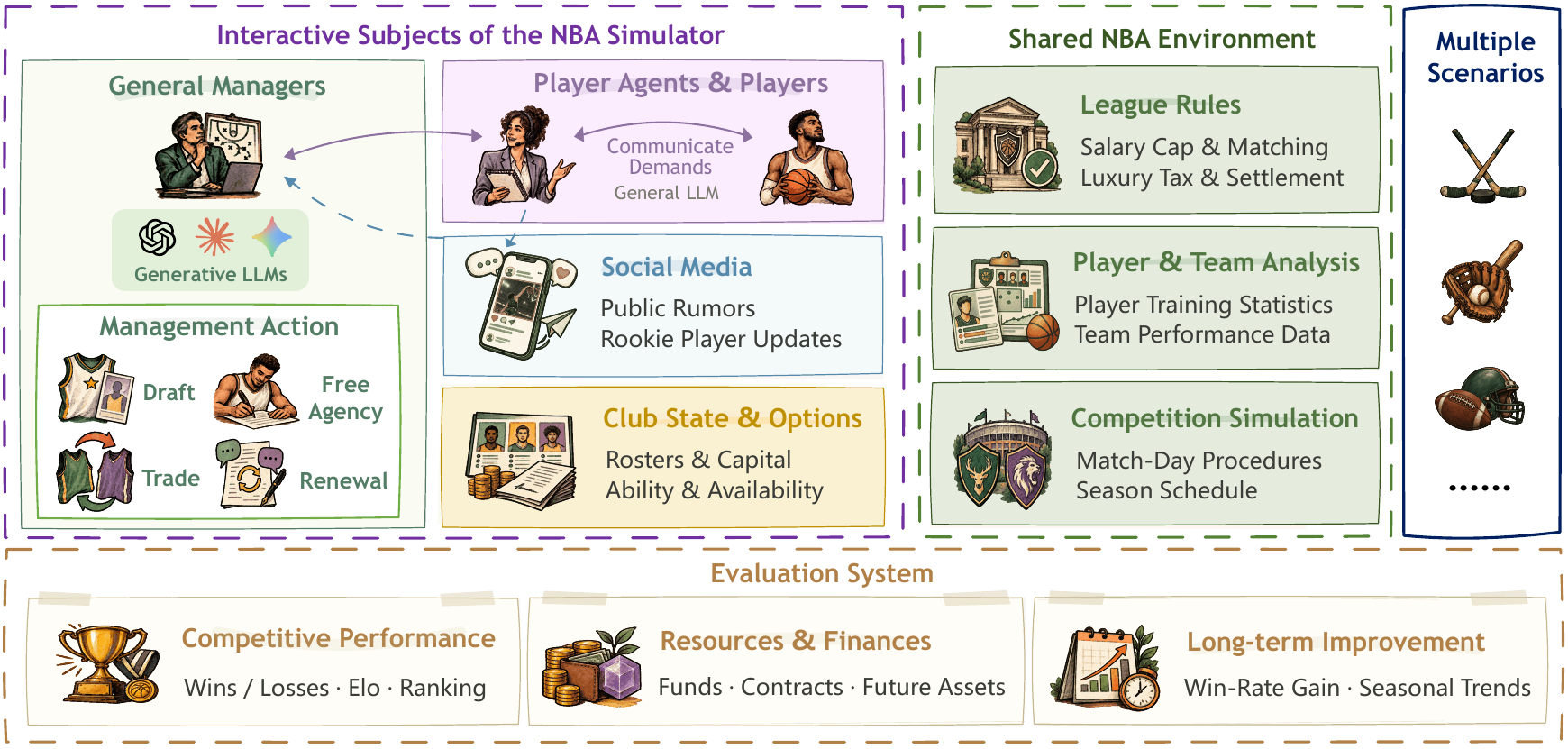}
    \caption{
    Overview of GM Bench, illustrated with the NBA environment and extended to hockey, baseball, and American football.
    LLM general managers interact with players and their agents within a shared simulator that enforces league and financial rules, while evaluation tracks competitive performance, resources and finances, and cross-season improvement.}
    \label{fig:fig2}
\end{figure*}

\subsection{State Dynamics and Outcomes}
Each GM observes team state, player demands, and public market information through $o_t^i$. Model $m_i$ for team $i$ conditions its actions on this observation and available history $h_t^i$, while the environment advances under joint action $\mathbf a_t$:
\begin{equation}
a_t^i\sim\pi_{m_i}(\cdot\mid o_t^i,h_t^i),\quad
s_{t+1}\sim P(\cdot\mid s_t,\mathbf a_t).
\end{equation}
Changes caused by actions persist and enter subsequent observations according to disclosure rules. The task combines repeated decisions, state dependencies across phases, and delayed feedback. Outcome vector $\mathbf U_i$ reports competitive performance, resource status, and long-term improvement, capturing cumulative returns and their resource costs.

\section{NBA Environment}
\label{sec:environment}
The NBA instance draws inspiration from ZenGM\footnote{\url{https://github.com/zengm-games/zengm}}, and models a NBA environment with league rules, LLM-based management, financial constraints, and role interactions.

\subsection{League Foundations}
\paragraph{Player Data and State.}
Initial rosters are constructed from real NBA player data based on player ratings. Growth and aging change ratings, while injuries affect availability and on-court performance. Game outcomes depend on player condition, roster composition, and random events, allowing the effects of roster adjustments to emerge throughout the season.

\paragraph{Core Rules and Financial Settings.}
The environment implements the following rules and financial mechanisms:
(1) \textbf{Game and season rules}, which define game procedures and regular-season and playoff schedules;
(2) \textbf{Salary cap}, which restricts payroll allocation and player acquisition;
(3) \textbf{Luxury tax}, which imposes additional costs when payroll exceeds the tax threshold;
(4) \textbf{Financial condition}, which updates cash balances using revenue, operating expenses, and luxury tax, with balances carried forward throughout the seasons.

\subsection{Market Interactions and GM Actions}
\label{sec:actions}
\paragraph{Player Demands and Market Signals.}
All players share an LLM that expresses trade demands based on team performance, playing roles, and contract status. Player representatives generate trade requests and public rumors. GMs combine these signals with roster needs to assess player value, conduct negotiation, and make offers.

\paragraph{Management Actions and Execution.}
GMs receive information on players, finances, and league rules and can perform four types of management actions:
(1) \textbf{Drafting}, selecting rookie players;
(2) \textbf{Signing}, signing free players;
(3) \textbf{Trading}, negotiating and revising offers involving players and draft picks with other GMs;
(4) \textbf{Contract renewal}, negotiating salaries with players after each season.

\section{General Sports Management}
\label{sec:protocol}
Differences in drafting, transfers, and registration across sports are implemented through the corresponding rules and game, while role interactions, intertemporal resource dependencies, and delayed feedback form a unified evaluation structure, as shown in Figure~\ref{fig:fig2}. GM Bench extends the NBA task to hockey, baseball, and American football to assess long-term management under different rules and roster needs.

\subsection{Evaluation Target: General Manager}
Each evaluated LLM serves as a team GM, allocating resources based on rosters, contracts, future assets, and market signals to improve competitive performance across seasons. Sports management follows a common feedback loop that links these decisions over time: GM actions change team states, shaping competitive and market outcomes that inform subsequent decisions. Within this loop, roster changes affect current strength, future payroll flexibility, and tradable assets, requiring GMs to coordinate immediate personnel decisions with long-term resource commitments.

\subsection{Environment Roles: Players and Agents}
Players and agents are both non-player characters (NPCs) rather than evaluation targets. They provide interaction partners and market signals by expressing individual demands and communicating trade opportunities, respectively.

\paragraph{Players: Expressing Personal Demands.}
All players share a general LLM with a fixed configuration. Based on individual circumstances, they express preferences over competitive prospects, playing roles, and compensation and participate in renewal negotiations. These preferences inform GM negotiation and resource allocation, while the simulator handles player development, injuries, and game performance.

\paragraph{Agents: Trade Requests and Public Rumors.}
Agents generate trade requests and public rumors for rostered players according to disclosure rules, turning individual circumstances into market signals. Private requests may remain undisclosed, while public rumors indicate potential trading opportunities. GMs weigh these signals against player condition, contract obligations, and roster needs when deciding whether to act.

\subsection{System Environment}
The system validates actions, advances games, and scores outcomes. Stochastic simulation and built-in management policies update league states and provide feedback for subsequent GM decisions.

\paragraph{League and Teams.}
The league checks contract renewals and financial compliance. Teams negotiate through LLMs, and agreements are validated before execution. Under the predefined season format, regular-season records determine standings and playoff eligibility, while playoffs decide the champion. At season end, the system settles revenue and expenses, opens drafting, signing, and renewal windows, and carries roster, contract, and financial states into the next season.

\paragraph{Coaches and Game Engine.}
Coaches and the game engine handle in-game decisions, translating roster configurations into outcomes shaped by roster, player, and randomness. GMs use this feedback to assess roster adjustments.

\begin{figure*}[!t]
\centering
\includegraphics[width=\textwidth]{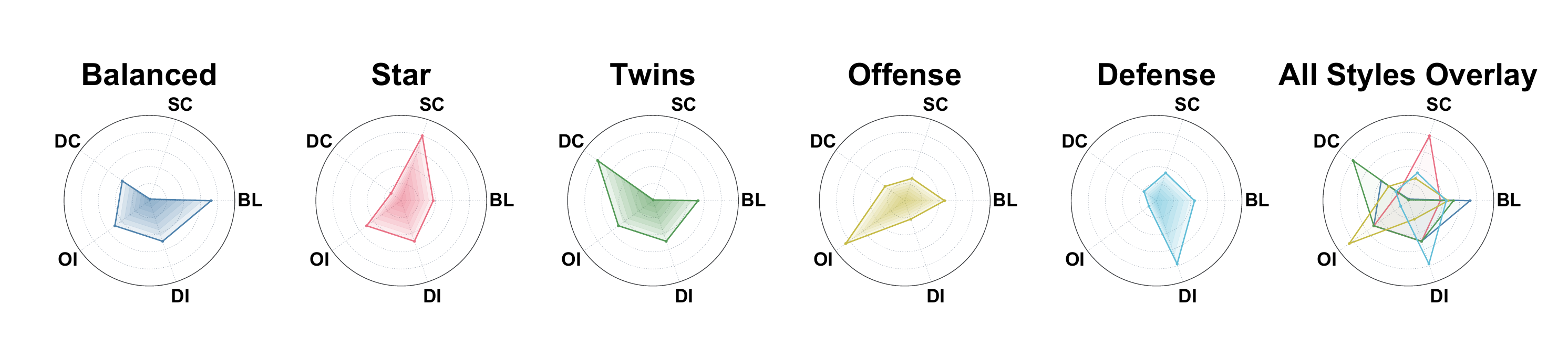}
\caption{\textbf{Schematic profiles of the five team types.} Balanced emphasizes an evenly distributed roster; Star centers on one standout player; Twins features two leading players; Offense prioritizes offensive ability; and Defense emphasizes defensive ability. The axes represent roster balance (\textbf{BL}), single-star contrast (\textbf{SC}), dual-core advantage (\textbf{DC}), offensive index (\textbf{OI}), and defensive index (\textbf{DI}). Larger radial values indicate stronger emphasis on the corresponding characteristic. The final panel overlays all five profiles for comparison.}
\label{fig:fig3}
\end{figure*}

\paragraph{Media and Market Information.}
The system controls the timing and visibility of player demands, public rumors, and media reports, while journalists and social media disseminate information. Draft windows provide prospect names, positions, ratings, and values, helping GMs identify market opportunities and assess roster fit.

\section{Data Sources and Evaluation}
\label{sec:benchmark}

We construct teams with comparable initial strength but distinct management needs from real-world player data to compare model strategies. Evaluation tracks competitive performance, resource retention, and long-term improvement, capturing cumulative decision returns and future investment capacity.

\subsection{Player Data and Team Initialization}
\label{sec:initialization}
For each of the four sports, we construct five roster types from real-world player data based on ratings, as shown in Figure~\ref{fig:fig3}, with comparable initial strength but distinct needs for bench depth, support for core players, and offensive or defensive improvement:
(1) \textbf{Balanced}, with relatively even strength across starters and reserves;
(2) \textbf{Star}, centered on one standout star with weaker supporting players;
(3) \textbf{Twins}, led by two core players;
(4) \textbf{Offense}, emphasizing shooting;
(5) \textbf{Defense}, emphasizing defensive ability.

\subsection{Evaluation Metrics}
\label{sec:metrics}
We report competitive performance, resource and financial status, and long-term improvement across four sports to characterize management returns and resource costs.

\paragraph{Competitive Performance.}
A common Elo protocol~\citep{elo1978rating} measures relative competitive performance within each league. All $N$ teams start each of $T$ seasons at $R_{i,s}^{(0)}=R_0=1500$, with no rating carryover. Games are processed chronologically, with $R_{i,s}^{(g)}$ denoting the rating of team $i$ after rated league game $g$ in season $s$. For teams $i$ and $j$, the pre-game expected score is
\begin{equation}
p_{i,s,g}
=\frac{1}{1+10^{(R_{j,s}^{(g-1)}-R_{i,s}^{(g-1)})/400}},
\end{equation}
with $p_{j,s,g}=1-p_{i,s,g}$. Both ratings are updated from the same pre-game values:
\begin{equation}
R_{k,s}^{(g)}
=R_{k,s}^{(g-1)}
+K\bigl(y_{k,s,g}-p_{k,s,g}\bigr),
\end{equation}
for $k\in\{i,j\}$. The observed score $y_{k,s,g}$ is $1$ for a win, $0$ for a loss, and $0.5$ for a draw where permitted. We fix $K=20$ for all teams and games; nonparticipants retain their ratings. Updates reward unexpected wins more strongly and preserve the league-wide total $\sum_{i=1}^{N}R_{i,s}^{(g)}=NR_0$.

With $M_s$ rated games in season $s$, we report mean season-end Elo:
\begin{equation}
E_{i,s}=R_{i,s}^{(M_s)},\qquad
\overline E_i=\frac{1}{T}\sum_{s=1}^{T}E_{i,s}.
\end{equation}
Higher $\overline E_i$ indicates stronger performance. We also report cumulative regular-season wins, losses, draws where applicable, and seasonal rank.

\paragraph{Resources and Finances.}
Terminal cash $C_i$ is the available balance after revenue and expenses are settled at the end of the management horizon, reflecting capacity for subsequent investment. Cash and competitive records jointly reveal the trade-off between resource retention and competitive returns. Active contracts and future assets can provide additional context on team resources.

\paragraph{Long-Term Improvement.}
Seasonal win-rate and cash track management performance. We define improvement as the difference between mean win rates in the final and initial $w$ seasons:
\begin{equation}
G_i=\frac{1}{w}\sum_{s=T-w+1}^{T}v_{i,s}
    -\frac{1}{w}\sum_{s=1}^{w}v_{i,s},
\end{equation}
where $v_{i,s}=W_{i,s}/(W_{i,s}+L_{i,s})$ is the regular-season win rate excluding draws, and $w$ is prespecified ($2w\leq T$). Positive and negative $G_i$ indicate improvement and decline, respectively. Together with $\overline E_i$, records, and terminal resources, $G_i$ characterizes long-term management outcomes.

\section{Experiments}
\subsection{Experimental Settings}
\paragraph{Models and Configuration.}
The evaluation comprises 31 LLMs from 12 families: OpenAI (7), Anthropic (4), Gemini, DeepSeek, and Qwen (3 each), Grok, GLM, MiMo, and Doubao (2 each), and Kimi, MiniMax, and HY (1 each); the details are shown in Appendix~\ref{app:model-identifiers}. The experiments span five seasons (2022-2026). We use official APIs with \texttt{high} reasoning effort where supported; more details are shown in Appendix~\ref{app:reproducibility}.

\paragraph{Baseline.}
We design \textbf{Native GM} as a rule-based management baseline: (1) Player valuation: players are valued using ability, age, potential, contracts, and injuries; (2) Strategy selection: team results and roster composition guide strategy selection, contention prioritizes current strength, while rebuilding favors young players, draft picks, and inexpensive contracts; (3) Contract management: signings prioritize high-value free agents under salary constraints; renewals consider contract costs and player willingness.

\subsection{Overall Results and Model Comparisons}
\paragraph{Overall Competitive Performance.}
In NBA setting, initially comparable teams show substantial divergence across seasons, as shown in Table~\ref{tab:nba-results}. Elo ranges from 1071.0 to 1977.3, with Claude Opus 5.5 ranking first and winning three championships, followed by GPT-6 Astra and GPT-6.1 Sol. Opus, Astra, and Sonnet collectively win all five championships. Regular-season wins span 35 to 47 in the first season and 8 to 76 in the final season, this widening spread captures the cumulative effects of management over time.

\begin{table*}[!t]
\centering
\begin{minipage}[t]{0.49\textwidth}\vspace{0pt}
\centering
\nbatablefont\fontsize{9}{10.8}\selectfont
\setlength{\tabcolsep}{0.3pt}
\renewcommand{\arraystretch}{1.07}
\setlength{\heavyrulewidth}{0.65pt}
\setlength{\lightrulewidth}{0.45pt}
\begin{tabularx}{\columnwidth}{@{}r@{\hspace{3pt}}>{\raggedright\arraybackslash}Xcccc@{}}
\toprule
\multicolumn{1}{@{}c}{\makebox[9pt][l]{\fontsize{10.5}{12.6}\selectfont\textbf{Rank}}} &
\multicolumn{1}{c}{\fontsize{10.5}{12.6}\selectfont\textbf{Model}} &
\multicolumn{1}{c}{\fontsize{10.5}{12.6}\selectfont\nbahl{nbaElo}{\textbf{Elo}}} &
\multicolumn{1}{c}{\fontsize{10.5}{12.6}\selectfont\nbahl{nbaWins}{\textbf{Wins}}} &
\multicolumn{1}{c}{\fontsize{10.5}{12.6}\selectfont\nbahl{nbaLosses}{\textbf{Losses}}} &
\multicolumn{1}{c@{}}{\fontsize{10.5}{12.6}\selectfont\nbahl{nbaCash}{\textbf{Cash}}} \\
\midrule
1 & Claude Opus 5.5\,\mbox{\goldmedal\kern0.5pt\goldmedal\kern0.5pt\goldmedal} & \nbahl{nbaEloBest}{\textbf{1977.3}} & \nbahl{nbaWinsBest}{\textbf{446}} & \nbahl{nbaLossesBest}{\textbf{66}} & 312.48 \\
2 & GPT-6 Astra\,\mbox{\goldmedal} & \underline{1907.7} & \underline{427} & \underline{89} & \underline{1006.62} \\
3 & GPT-6.1 Sol & 1838.7 & 400 & 101 & \nbahl{nbaCashBest}{\textbf{1152.33}} \\
4 & Claude Sonnet 5.5\,\mbox{\goldmedal} & 1813.3 & 385 & 98 & 308.05 \\
5 & Claude Fable 5.1 & 1794.4 & 377 & 106 & 419.07 \\
6 & Grok 4.7 & 1784.3 & 347 & 121 & 585.74 \\
7 & Gemini 3.1 Pro Preview & 1753.1 & 334 & 131 & \nbahl{nbaCashNegative}{-21.63} \\
8 & Grok 4.6 & 1713.6 & 316 & 150 & 443.52 \\
9 & Kimi K3 & 1698.8 & 308 & 130 & 868.23 \\
10 & DeepSeek-V4-Pro & 1666.3 & 275 & 161 & 255.43 \\
11 & GLM-5.3 & 1652.6 & 279 & 161 & 188.42 \\
12 & Qwen3.8-Max & 1602.5 & 250 & 189 & 292.10 \\
13 & GPT-6 Luna & 1582.6 & 249 & 190 & \nbahl{nbaCashNegative}{-41.43} \\
14 & {\setlength{\fboxsep}{1pt}\setlength{\fboxrule}{0.5pt}\fbox{Native GM strategy}} & 1575.0 & 240 & 196 & 96.78 \\
15 & Gemini 3.8 Flash & 1525.7 & 220 & 217 & 111.40 \\
16 & GPT-5.6 Terra & 1499.0 & 196 & 232 & 527.42 \\
17 & MiniMax M3 & 1451.8 & 193 & 226 & 106.26 \\
18 & DeepSeek-V4.1-Flash & 1418.4 & 178 & 232 & 294.00 \\
19 & Qwen3.8-Flash & 1388.5 & 154 & 256 & 428.19 \\
20 & Qwen3.8-27B & 1361.6 & 154 & 256 & 460.73 \\
21 & Doubao-Seed-2.1-Pro & 1331.0 & 133 & 277 & 335.05 \\
22 & GLM-5.3-Flash & 1305.8 & 117 & 293 & 581.30 \\
23 & MiMo-V2.6-Pro & 1275.4 & 105 & 305 & 925.41 \\
24 & Gemini 3.5 Flash-Lite & 1255.2 & 106 & 304 & 65.99 \\
25 & GPT-4o & 1206.4 & 91 & 319 & 360.08 \\
26 & DeepSeek-R1 & 1167.2 & 81 & 329 & 167.76 \\
27 & Claude Haiku 4.5 & 1150.6 & 61 & 349 & 244.02 \\
28 & Hy4 preview & 1124.7 & 53 & 357 & 81.31 \\
29 & MiMo-V2.6-Flash & 1107.2 & 51 & 359 & \nbahl{nbaCashNegative}{-40.19} \\
30 & Doubao-Seed-2.1-Turbo & 1071.0 & 42 & 368 & 354.07 \\
\bottomrule
\end{tabularx}
\caption{NBA results on GM Bench. Each \goldmedal{} denotes one championship. Best values are bold and highlighted; second-best values are underlined. Negative cash is highlighted in red.}
\label{tab:nba-results}
\end{minipage}\hfill
\begin{minipage}[t]{0.49\textwidth}\vspace{0pt}
\centering
\begingroup
\nbatablefont\fontsize{9}{10.8}\selectfont
\setlength{\tabcolsep}{0.3pt}
\renewcommand{\arraystretch}{1.07}
\setlength{\heavyrulewidth}{0.65pt}
\setlength{\lightrulewidth}{0.45pt}
\definecolor{tTwoBlue}{RGB}{205,228,247}
\definecolor{tTwoPurple}{RGB}{226,217,243}
\definecolor{tTwoPink}{RGB}{243,208,226}
\definecolor{tTwoOrange}{RGB}{250,223,197}
\definecolor{tTwoGreen}{RGB}{211,234,218}
\begin{tabularx}{\columnwidth}{@{}r@{\hspace{3pt}}>{\raggedright\arraybackslash}Xcccccc@{}}
\toprule
\multicolumn{1}{@{}c}{\makebox[9pt][l]{\fontsize{10.5}{12.6}\selectfont\textbf{Rank}}} & \multicolumn{1}{c}{\fontsize{10.5}{12.6}\selectfont\textbf{Model}} & {\fontsize{10.5}{12.6}\selectfont\nbahl{tTwoBlue}{\textbf{22}}} & {\fontsize{10.5}{12.6}\selectfont\nbahl{tTwoPurple}{\textbf{23}}} & {\fontsize{10.5}{12.6}\selectfont\nbahl{tTwoPink}{\textbf{24}}} & {\fontsize{10.5}{12.6}\selectfont\nbahl{tTwoOrange}{\textbf{25}}} & {\fontsize{10.5}{12.6}\selectfont\nbahl{tTwoGreen}{\textbf{26}}} & {\fontsize{10.5}{12.6}\selectfont\boldmath$G_i$} \\
\midrule
1 & \mbox{Claude Opus 5.5} & 44 & \nbahl{tTwoPurple}{\textbf{54}} & \underline{65} & \nbahl{tTwoOrange}{\textbf{74}} & \nbahl{tTwoGreen}{\textbf{76}} & \textbf{+31.71} \\
2 & \mbox{GPT-6 Astra} & 45 & 50 & 61 & \underline{69} & \underline{73} & \textbf{+28.66} \\
3 & \mbox{GPT-6.1 Sol} & 45 & 49 & 60 & 67 & 71 & \textbf{+26.83} \\
4 & \mbox{Claude Sonnet 5.5} & 43 & \nbahl{tTwoPurple}{\textbf{54}} & \nbahl{tTwoPink}{\textbf{75}} & 65 & 56 & \textbf{+14.63} \\
5 & \mbox{Claude Fable 5.1} & 38 & 49 & 64 & 67 & 66 & \textbf{+28.05} \\
6 & \mbox{Grok 4.7} & 36 & 46 & 50 & 59 & 63 & \textbf{+24.39} \\
7 & \mbox{Gemini 3.1 Pro Preview} & 42 & 51 & 64 & 57 & 46 & \textbf{+6.10} \\
8 & \mbox{Grok 4.6} & 38 & 46 & 49 & 55 & 57 & \textbf{+17.07} \\
9 & \mbox{Kimi K3} & 42 & 46 & 55 & 58 & 62 & \textbf{+19.51} \\
10 & \mbox{DeepSeek-V4-Pro} & 36 & 44 & 46 & 53 & 56 & \textbf{+17.68} \\
11 & \mbox{GLM-5.3} & \nbahl{tTwoBlue}{\textbf{47}} & \underline{52} & 53 & 54 & 57 & \textbf{+7.32} \\
12 & \mbox{Qwen3.8-Max} & 41 & 42 & 48 & 48 & 52 & \textbf{+10.37} \\
13 & \mbox{GPT-6 Luna} & 37 & 48 & 54 & 46 & 38 & \textbf{\ensuremath{-}0.61} \\
14 & {\setlength{\fboxsep}{1pt}\setlength{\fboxrule}{0.5pt}\fbox{Native GM strategy}} & 39 & 44 & 46 & 48 & 49 & \textbf{+8.54} \\
15 & \mbox{Gemini 3.8 Flash} & 40 & 40 & 41 & 42 & 43 & \textbf{+3.05} \\
16 & \mbox{GPT-5.6 Terra} & 44 & 38 & 38 & 38 & 38 & \textbf{\ensuremath{-}3.66} \\
17 & \mbox{MiniMax M3} & 39 & 38 & 38 & 38 & 41 & \textbf{+1.22} \\
18 & \mbox{DeepSeek-V4.1-Flash} & \nbahl{tTwoBlue}{\textbf{47}} & 38 & 37 & 36 & 38 & \textbf{\ensuremath{-}6.71} \\
19 & \mbox{Qwen3.8-Flash} & 37 & 35 & 34 & 32 & 32 & \textbf{\ensuremath{-}4.88} \\
20 & \mbox{Qwen3.8-27B} & 35 & 33 & 32 & 32 & 32 & \textbf{\ensuremath{-}2.44} \\
21 & \mbox{Doubao-Seed-2.1-Pro} & 35 & 32 & 27 & 27 & 27 & \textbf{\ensuremath{-}7.93} \\
22 & \mbox{GLM-5.3-Flash} & 40 & 36 & 25 & 24 & 27 & \textbf{\ensuremath{-}15.24} \\
23 & \mbox{MiMo-V2.6-Pro} & 40 & 37 & 23 & 22 & 22 & \textbf{\ensuremath{-}20.12} \\
24 & \mbox{Gemini 3.5 Flash-Lite} & 42 & 34 & 22 & 24 & 23 & \textbf{\ensuremath{-}17.68} \\
25 & \mbox{GPT-4o} & \underline{46} & 35 & 29 & 23 & 19 & \textbf{\ensuremath{-}23.78} \\
26 & \mbox{DeepSeek-R1} & 39 & 32 & 18 & 19 & 17 & \textbf{\ensuremath{-}21.34} \\
27 & \mbox{Claude Haiku 4.5} & 41 & 32 & 18 & 14 & 15 & \textbf{\ensuremath{-}26.83} \\
28 & \mbox{HY4 Preview} & \underline{46} & 32 & 18 & 13 & 12 & \textbf{\ensuremath{-}32.32} \\
29 & \mbox{MiMo-V2.6-Flash} & 43 & 31 & 22 & 14 & 8 & \textbf{\ensuremath{-}31.71} \\
30 & \mbox{Doubao-Seed-2.1-Turbo} & 43 & 32 & 18 & 12 & 14 & \textbf{\ensuremath{-}29.88} \\
\bottomrule
\end{tabularx}
\endgroup
\caption{NBA regular-season wins and long-term improvement. Rank follows Table~\ref{tab:nba-results}. $G_i$ is the change in average win rate from 2022--2023 to 2025--2026 ($w=2$), expressed in percentage points.}
\label{tab:nba-long-term-improvement}
\end{minipage}
\end{table*}

\paragraph{Native Policy as a Strategic Reference.}
The native policy ranks 14th in the NBA setting, with regular-season wins rising from 39 to 49. Thirteen LLMs rank above it and 16 below, including Gemini 3.8 Flash, GPT-5.6 Terra, and DeepSeek-V4.1-Flash. These results highlight room for improvement in the long-term management capabilities of many current LLMs.

\paragraph{Strategic Capacity Across Model Generations.}
Long-term management performance varies substantially across model generations, as shown in Table~\ref{tab:nba-results}. Within families, GPT-6 Astra and GPT-6.1 Sol outperform GPT-4o, DeepSeek-V4-Pro exceeds DeepSeek-R1. Differences also emerge across tiers within the same generation, with Qwen3.8-Max exceeding Qwen3.8-Flash. However, older models can outperform newer releases both within and across families: GPT-4o exceeds Hy4 Preview and MiMo-V2.6-Flash, while Gemini 3.1 Pro Preview outperforms Gemini 3.8 Flash.

\subsection{Resource Allocation Across Seasons}
\paragraph{Star Acquisition and Competitive Returns.}
Trading observations and terminal cash balances reveal contrasting investment styles, as shown in Table~\ref{tab:nba-results}: Claude 5 GMs actively trade for championship contention, whereas GPT-6 GMs favor cautious acquisitions. Opus acquires stars and wins three championships, while Astra and Sol preserve larger cash buffers for future roster adjustments. Grok 4.7 achieves stronger competitive returns with less remaining cash, whereas Kimi K3 combines sustained improvement with greater financial flexibility.  These comparisons highlight competitive returns and future investment capacity as joint criteria for assessing star acquisitions.

\paragraph{Championship Peaks and Roster Continuity.}
Peak success and sustained improvement can diverge, as shown in Table~\ref{tab:nba-long-term-improvement}. Sonnet wins a championship and peaks at 75 regular-season wins in 2024 before falling to 65 and 56. Sol improves steadily from 45 to 71, achieving a larger $G_i$. Fable records 67 and 66 wins in the final two seasons and achieving $G_i=28.05$ percentage points without a championship. Renewal observations show that Sonnet faces unaffordable salary demands and loses valuable players, highlighting the need to coordinate acquisition and retention.

\paragraph{Financial Feedback and Investment Capacity.}
Cash balances complement competitive results as a measure of investment capacity. Astra and Sol retain 1006.62 and 1152.33, respectively, compared with 312.48 for Opus. MiMo-V2.6-Pro retains 925.41 in cash but ranks 23rd, with regular-season wins falling from 40 to 22, showing high balances may coexist with weak competitive returns. Cash deficits occur at different competitive levels: Gemini 3.1 Pro Preview (rank 7, $-21.63$), GPT-6 Luna (rank 13, $-41.43$), and MiMo-V2.6-Flash (rank 29, $-40.19$). These management cases illustrate linked financial pressures: salary commitments tie up future revenue, while weak results reduce revenue and constrain reinvestment.

\FloatBarrier

\begin{table*}[!t]
\centering
\begingroup
\fontsize{7.4}{9.6}\selectfont
\setlength{\tabcolsep}{0.8pt}
\renewcommand{\arraystretch}{1.12}
\begin{minipage}[t]{0.322\textwidth}\vspace{0pt}
\centering
{\fontsize{9}{10.8}\selectfont\textbf{(a) American football}}\par\smallskip
\begin{tabularx}{\linewidth}{@{}r >{\raggedright\arraybackslash}X r r@{}}
\toprule
\multicolumn{1}{@{}c}{\fontsize{9}{10.8}\selectfont\textbf{Rank}} & \multicolumn{1}{c}{\fontsize{9}{10.8}\selectfont\textbf{Model}} & {\fontsize{9}{10.8}\selectfont\nbahl{nbaElo}{\textbf{Elo}}} & {\fontsize{9}{10.8}\selectfont\nbahl{nbaCash}{\textbf{Cash}}} \\
\midrule
1 & \mbox{Claude Opus 5.5} & \nbahl{nbaEloBest}{\textbf{1870.1}} & 228.19 \\
2 & \mbox{Claude Sonnet 5.5} & \underline{1761.6} & 78.00 \\
3 & \mbox{GPT-6.1 Sol} & 1760.0 & 205.34 \\
4 & \mbox{Kimi K3} & 1750.0 & 86.48 \\
5 & \mbox{GPT-6 Astra} & 1745.0 & 163.94 \\
6 & \mbox{Claude Fable 5.1} & 1736.3 & 236.92 \\
7 & \mbox{Grok 4.7} & 1733.0 & \nbahl{nbaCashBest}{\textbf{422.70}} \\
8 & \mbox{Gemini 3.1 Pro Preview} & 1714.4 & 210.74 \\
9 & \mbox{DeepSeek-V4-Pro} & 1714.3 & 117.67 \\
10 & \mbox{GLM-5.3} & 1700.0 & 233.54 \\
11 & \mbox{Grok 4.6} & 1663.8 & 215.06 \\
12 & \mbox{GPT-6 Luna} & 1644.4 & 178.54 \\
13 & \mbox{Gemini 3.8 Flash} & 1630.4 & 191.05 \\
14 & \mbox{Qwen3.8-Max} & 1559.7 & 271.95 \\
15 & \mbox{GPT-5.6 Terra} & 1547.8 & 102.12 \\
16 & \mbox{DeepSeek-V4.1-Flash} & 1518.7 & 93.57 \\
17 & \mbox{MiniMax M3} & 1468.9 & \nbahl{nbaCashNegative}{-67.53} \\
18 & \mbox{Qwen3.8-Flash} & 1401.8 & 161.60 \\
19 & {\setlength{\fboxsep}{1pt}\setlength{\fboxrule}{0.4pt}\fbox{Native GM strategy}} & 1386.1 & 262.38 \\
20 & \mbox{MiMo-V2.6-Pro} & 1385.0 & 74.50 \\
21 & \mbox{Qwen3.8-27B} & 1365.1 & \underline{391.39} \\
22 & \mbox{GPT-5.4 Mini} & 1352.7 & 196.24 \\
23 & \mbox{Gemini 3.5 Flash-Lite} & 1320.5 & 123.38 \\
24 & \mbox{Doubao-Seed-2.1-Pro} & 1315.4 & 37.75 \\
25 & \mbox{GLM-5.3-Flash} & 1310.2 & 248.22 \\
26 & \mbox{DeepSeek-R1} & 1258.2 & 173.47 \\
27 & \mbox{MiMo-V2.6-Flash} & 1245.0 & 42.79 \\
28 & \mbox{GPT-5.4 Nano} & 1220.8 & 211.02 \\
29 & \mbox{Hy4 preview} & 1210.6 & 126.84 \\
30 & \mbox{GPT-4o} & 1193.3 & 210.78 \\
31 & \mbox{Doubao-Seed-2.1-Turbo} & 1163.5 & 169.85 \\
32 & \mbox{Claude Haiku 4.5} & 1105.3 & 221.24 \\
\bottomrule
\end{tabularx}
\end{minipage}
\hfill
\begin{minipage}[t]{0.322\textwidth}\vspace{0pt}
\centering
{\fontsize{9}{10.8}\selectfont\textbf{(b) Ice hockey}}\par\smallskip
\begin{tabularx}{\linewidth}{@{}r >{\raggedright\arraybackslash}X r r@{}}
\toprule
\multicolumn{1}{@{}c}{\fontsize{9}{10.8}\selectfont\textbf{Rank}} & \multicolumn{1}{c}{\fontsize{9}{10.8}\selectfont\textbf{Model}} & {\fontsize{9}{10.8}\selectfont\nbahl{nbaElo}{\textbf{Elo}}} & {\fontsize{9}{10.8}\selectfont\nbahl{nbaCash}{\textbf{Cash}}} \\
\midrule
1 & \mbox{Claude Opus 5.5} & \nbahl{nbaEloBest}{\textbf{1855.4}} & 63.10 \\
2 & \mbox{GPT-6.1 Sol} & \underline{1824.0} & 59.27 \\
3 & \mbox{Claude Sonnet 5.5} & 1768.8 & 40.08 \\
4 & \mbox{Gemini 3.1 Pro Preview} & 1754.7 & 138.45 \\
5 & \mbox{GPT-6 Astra} & 1750.0 & 0.18 \\
6 & \mbox{Claude Fable 5.1} & 1747.8 & 116.69 \\
7 & \mbox{Grok 4.7} & 1739.5 & 44.86 \\
8 & \mbox{Kimi K3} & 1665.4 & 267.86 \\
9 & \mbox{Grok 4.6} & 1617.6 & \underline{320.32} \\
10 & \mbox{DeepSeek-V4-Pro} & 1591.9 & 176.44 \\
11 & {\setlength{\fboxsep}{1pt}\setlength{\fboxrule}{0.4pt}\fbox{Native GM strategy}} & 1572.7 & \nbahl{nbaCashNegative}{-18.21} \\
12 & \mbox{Qwen3.8-Max} & 1569.3 & 49.08 \\
13 & \mbox{Qwen3.8-Flash} & 1557.6 & 116.02 \\
14 & \mbox{Gemini 3.8 Flash} & 1545.9 & 6.45 \\
15 & \mbox{GPT-5.4 Mini} & 1544.4 & 10.93 \\
16 & \mbox{GPT-5.6 Terra} & 1429.3 & \nbahl{nbaCashNegative}{-43.07} \\
17 & \mbox{DeepSeek-V4.1-Flash} & 1415.2 & 6.97 \\
18 & \mbox{GLM-5.3} & 1391.0 & 75.21 \\
19 & \mbox{Qwen3.8-27B} & 1390.2 & 217.59 \\
20 & \mbox{MiniMax M3} & 1380.2 & \nbahl{nbaCashNegative}{-58.10} \\
21 & \mbox{GPT-5.4 Nano} & 1347.0 & 69.68 \\
22 & \mbox{GPT-6 Luna} & 1333.0 & 51.33 \\
23 & \mbox{GLM-5.3-Flash} & 1317.6 & 204.81 \\
24 & \mbox{Doubao-Seed-2.1-Pro} & 1290.5 & \nbahl{nbaCashNegative}{-39.41} \\
25 & \mbox{GPT-4o} & 1251.1 & 17.16 \\
26 & \mbox{MiMo-V2.6-Pro} & 1237.3 & \nbahl{nbaCashNegative}{-10.39} \\
27 & \mbox{Claude Haiku 4.5} & 1235.6 & \nbahl{nbaCashBest}{\textbf{321.47}} \\
28 & \mbox{DeepSeek-R1} & 1191.0 & 80.67 \\
29 & \mbox{Doubao-Seed-2.1-Turbo} & 1186.8 & 119.83 \\
30 & \mbox{Hy4 preview} & 1166.3 & \nbahl{nbaCashNegative}{-11.50} \\
31 & \mbox{MiMo-V2.6-Flash} & 1157.1 & \nbahl{nbaCashNegative}{-5.79} \\
32 & \mbox{Gemini 3.5 Flash-Lite} & 1118.8 & \nbahl{nbaCashNegative}{-36.56} \\
\bottomrule
\end{tabularx}
\end{minipage}
\hfill
\begin{minipage}[t]{0.322\textwidth}\vspace{0pt}
\centering
{\fontsize{9}{10.8}\selectfont\textbf{(c) Baseball}}\par\smallskip
\begin{tabularx}{\linewidth}{@{}r >{\raggedright\arraybackslash}X r r@{}}
\toprule
\multicolumn{1}{@{}c}{\fontsize{9}{10.8}\selectfont\textbf{Rank}} & \multicolumn{1}{c}{\fontsize{9}{10.8}\selectfont\textbf{Model}} & {\fontsize{9}{10.8}\selectfont\nbahl{nbaElo}{\textbf{Elo}}} & {\fontsize{9}{10.8}\selectfont\nbahl{nbaCash}{\textbf{Cash}}} \\
\midrule
1 & \mbox{GPT-6 Astra} & \nbahl{nbaEloBest}{\textbf{1895.5}} & 203.05 \\
2 & \mbox{Claude Opus 5.5} & \underline{1860.4} & \nbahl{nbaCashNegative}{-33.64} \\
3 & \mbox{GPT-6.1 Sol} & 1780.0 & 149.24 \\
4 & \mbox{Qwen3.8-Max} & 1776.2 & 386.26 \\
5 & \mbox{Claude Sonnet 5.5} & 1769.4 & \nbahl{nbaCashBest}{\textbf{527.73}} \\
6 & \mbox{Grok 4.7} & 1747.2 & \underline{447.90} \\
7 & \mbox{Claude Fable 5.1} & 1737.1 & 32.27 \\
8 & {\setlength{\fboxsep}{1pt}\setlength{\fboxrule}{0.4pt}\fbox{Native GM strategy}} & 1727.0 & 406.52 \\
9 & \mbox{Gemini 3.1 Pro Preview} & 1719.6 & 373.27 \\
10 & \mbox{GPT-6 Luna} & 1707.3 & 323.11 \\
11 & \mbox{Grok 4.6} & 1695.1 & 287.21 \\
12 & \mbox{GLM-5.3} & 1666.9 & 264.28 \\
13 & \mbox{DeepSeek-V4-Pro} & 1623.5 & 242.39 \\
14 & \mbox{Kimi K3} & 1616.6 & 357.41 \\
15 & \mbox{MiniMax M3} & 1540.1 & \nbahl{nbaCashNegative}{-41.12} \\
16 & \mbox{Doubao-Seed-2.1-Pro} & 1505.6 & \nbahl{nbaCashNegative}{-686.43} \\
17 & \mbox{GLM-5.3-Flash} & 1476.3 & 42.31 \\
18 & \mbox{Qwen3.8-27B} & 1441.3 & 151.91 \\
19 & \mbox{GPT-5.6 Terra} & 1441.1 & 43.03 \\
20 & \mbox{MiMo-V2.6-Pro} & 1380.5 & \nbahl{nbaCashNegative}{-609.77} \\
21 & \mbox{Gemini 3.8 Flash} & 1378.0 & \nbahl{nbaCashNegative}{-14.77} \\
22 & \mbox{DeepSeek-V4.1-Flash} & 1350.9 & 21.55 \\
23 & \mbox{Gemini 3.5 Flash-Lite} & 1346.8 & \nbahl{nbaCashNegative}{-39.27} \\
24 & \mbox{Qwen3.8-Flash} & 1302.4 & 106.73 \\
25 & \mbox{Claude Haiku 4.5} & 1263.8 & 172.99 \\
26 & \mbox{GPT-4o} & 1256.5 & 60.17 \\
27 & \mbox{DeepSeek-R1} & 1237.9 & 165.93 \\
28 & \mbox{Hy4 preview} & 1216.5 & 206.73 \\
29 & \mbox{Doubao-Seed-2.1-Turbo} & 1146.9 & \nbahl{nbaCashNegative}{-22.45} \\
30 & \mbox{MiMo-V2.6-Flash} & 1084.9 & 53.63 \\
 & & & \\
 & & & \\
\bottomrule
\end{tabularx}
\end{minipage}
\endgroup
\caption{Elo rankings and terminal cash in (a) American football, (b) ice hockey, and (c) baseball. The highest value in each column and sport is bold and highlighted; the second-highest is underlined. Negative cash is highlighted in red. The native GM strategy is boxed.}
\label{tab:additional-sports-results}
\end{table*}

\subsection{Results Across Additional Sports}
\paragraph{American Football.}
Claude Opus 5.5 leads American football, matching its first-place NBA result, while Kimi K3 ranks fourth compared with ninth in the NBA, as shown in Table~\ref{tab:nba-long-term-improvement}. The six highest-ranked NBA models remain in the top seven in all three additional sports, showing consistent competitive strength across settings. The native GM strategy ranks 19th, below its NBA placement of 14th. Grok 4.7 combines seventh place with the highest terminal cash, showing that competitive leadership and cash retention remain distinct across settings.

\paragraph{Ice Hockey.}
Opus again leads ice hockey, followed by GPT-6.1 Sol, preserving their strong NBA standings. Financial outcomes transfer less uniformly: Astra, one of the two NBA cash leaders, ranks fifth but retains only 0.18, whereas Haiku holds the largest balance despite ranking 27th. The native GM strategy ranks 11th, above its NBA and American football placements, but finishes with negative cash. Across evaluated teams, terminal deficits occur in 8 of 32 hockey teams and 7 of 30 baseball teams, compared with 1 of 32 in American football and 3 of 30 in the NBA. These results show that stronger competitive performance can coexist with financial strain.

\paragraph{Baseball.}
GPT-6 Astra leads baseball, compared with second in the NBA and fifth in both American football and ice hockey. Qwen3.8-Max and Kimi K3 reverse their relative positions: Qwen ranks 14th in American football and fourth in baseball, while Kimi ranks fourth and 14th, respectively. Opus, which leads the other three sports, ranks second but ends with negative cash. Its aggressive investment leads to bankruptcy late in the evaluation, followed by player departures and a series of losses. Sonnet combines fifth place with the highest cash balance, further illustrating how financial rankings shift across sports. The native GM strategy ranks eighth. Together, these experiments assess how LLMs adapt resource allocation across sports to sustain competitive performance and future investment capacity.

\section{Related Work}
\paragraph{Interactive Evaluation.}
Interactive benchmarks evaluate sequential actions and coordination in executable environments~\citep{agentboard,multiagentbench}. WebArena, AppWorld, and TheAgentCompany cover web, application, and workplace workflows~\citep{webarena,appworld,theagentcompany}, while BrowserGym unifies web-agent evaluation~\citep{browsergym}. $\tau^2$-bench tests agent--user collaboration in a shared environment~\citep{tau2bench}, and BALROG evaluates planning and exploration across games~\citep{balrog}.

\paragraph{Long-Term Management and Competition.}
Recent benchmarks examine the long-term effects of management decisions. Vending-Bench, EcoGym, and EnterpriseArena evaluate sustained business operation and resource allocation~\citep{vendingbench,ecogym,enterprisearena}. Market-Bench, Vending-Bench Arena, and CoffeeBench study procurement, pricing, and trading in multi-agent markets~\citep{marketbench,coffeebench}. E-Commerce Bench grounds year-long operations in real product and supplier data, linking repeated supplier negotiations, cash-flow management, and long-term returns~\citep{ecommercebench}.

\paragraph{Sports Management.}
Sports management links player development, contract constraints, and performance across seasons. FM-Bench combines long-term management, model competition, and equal-budget drafting in a synthetic football world~\citep{fmbench}. FromPitch2Board uses paired seeds and controlled configurations to examine models, agent scaffolds, management roles, and evaluation horizons~\citep{frompitch2board}. GM Bench evaluates cross-season resource allocation under competition across four sports, using real player data, roster types of comparable strength, and a shared LLM for player demands and negotiation.

\section{Conclusion}
We introduced \textbf{GM Bench}, an interactive benchmark for long-horizon sports management. A shared task across four sports connects competition among LLM general managers, player interactions, and persistent resource constraints to outcomes across seasons. Experiments reveal distinct balances between championship pursuit, financial flexibility, and roster continuity. Renewal trajectories link competitive success to subsequent retention demands, highlighting anticipation of future commitments as a key element of sustained improvement. By jointly tracking competitive performance, cash reserves, and seasonal trajectories, GM Bench makes cumulative decision effects observable and evaluates the ability to sustain progress under competition.

\section*{Limitations}

\paragraph{League Configuration.}
Each sport is instantiated through a specified simulator and league configuration. The unified task structure supports evaluation across four sports, while contract systems, roster constraints, financial rules, and competition formats remain specific to each implementation. Extending the benchmark to other leagues requires corresponding rule and interface adaptations.

\paragraph{Observation Horizon.}
Decisions are assessed over a finite number of seasons, capturing cumulative competitive and financial effects within the evaluation window. Full player careers and distant future assets may extend beyond this horizon. Longer evaluations would help characterize delayed player development, aging, and the outcomes of commitments that mature after the evaluation ends.

\paragraph{Evaluation Cost.}
Multi-season evaluation with interacting LLM roles requires repeated inference and simulation. Expanding coverage to additional models, random seeds, and league configurations increases computational cost and evaluation time. More efficient execution and parallelization of independent runs could facilitate broader experimental coverage.

\section*{Ethical Considerations}

GM Bench uses real-player data to initialize simulated states. Player demands, negotiations, and market rumors are generated as fictional scenario content and do not represent the actual views, intentions, or conduct of the individuals portrayed. We interpret these interactions within their simulated roles and restrict evaluation conclusions to model behavior. All management actions are executed within the simulator, and benchmark outputs are not intended to guide real-world recruitment, contract negotiations, or financial decisions.

We comply with applicable source licenses and attribution requirements when reusing data and software. We document the provenance and intended use of datasets, prompts, and trajectories, and exclude private personal information and confidential content from shared materials. Our paper distinguishes source player attributes from generated interactions, avoiding the presentation of synthetic statements as factual reports about real individuals.


\bibliography{custom}
\bibliographystyle{acl_natbib}

\clearpage

\appendix

\section{Benchmark Comparison}
\label{sec:Benchmark_Comparison}

Table~\ref{tab:benchmarks} compares representative benchmarks along four design dimensions:
\textbf{Demand negotiation} incorporates demands from LLM-driven personnel into resource allocation;
\textbf{Needs adaptation} tests strategy adjustment across teams with comparable initial strength but different configuration needs;
\textbf{Intertemporal diagnostics} traces long-term action benefits and subsequent resource constraints through controlled comparisons or state replay;
\textbf{Institutional adaptation} evaluates strategies across sports rules and resources within a unified management task.
We also report the number of interacting agent roles in each environment.

\begin{table}[H]
\begin{minipage}{\linewidth}
\centering
\begingroup
\footnotesize
\setlength{\tabcolsep}{3.5pt}
\renewcommand{\arraystretch}{1.10}
\renewcommand{\tabularxcolumn}[1]{m{#1}}

\newcommand{\benchentry}[2]{%
  \mbox{#1~{\color{black!65}\citep{#2}}}%
}

\begin{tabularx}{\linewidth}{@{}
  >{\raggedright\arraybackslash}m{0.23\linewidth}
  >{\raggedright\arraybackslash}m{0.29\linewidth}
  *{4}{>{\centering\arraybackslash}X}
@{}}
\toprule
\rowcolor{BenchHeader}
\textbf{Benchmark}
& \textbf{\shortstack[l]{Interacting agents\\per environment}}
& \textbf{\shortstack{Demand\\negotiation}}
& \textbf{\shortstack{Needs\\adaptation}}
& \textbf{\shortstack{Long-term\\diagnostics}}
& \textbf{\shortstack{Institutional\\adaptation}} \\
\midrule

\benchentry{EcoGym}{ecogym}
& \textbf{1} management agent
& $\times$ & $\times$ & $\triangle$ & $\times$ \\
\addlinespace[2pt]

\benchentry{EnterpriseArena}{enterprisearena}
& \textbf{1} CFO agent
& $\times$ & $\times$ & $\triangle$ & $\times$ \\
\addlinespace[2pt]

\benchentry{FM-Bench Arena}{fmbench}
& \textbf{16} managers (15 LLM + 1 scripted)
& $\times$ & $\times$ & $\checkmark$ & $\times$ \\
\addlinespace[2pt]

\benchentry{Market-Bench}{marketbench}
& \textbf{20} LLM retailers
& $\times$ & $\times$ & $\times$ & $\times$ \\
\addlinespace[2pt]

\benchentry{CoffeeBench}{coffeebench}
& \textbf{6} LLM firms (1 evaluated + 5 reference)
& $\times$ & $\times$ & $\triangle$ & $\times$ \\
\addlinespace[2pt]

\benchentry{E-Commerce Bench}{ecommercebench}
& \textbf{1} manager + up to \textbf{576} supplier NPCs
& $\times$ & $\times$ & $\times$ & $\times$ \\

\midrule
\rowcolor{BenchHighlight}
\textbf{GM Bench}
& \textbf{30} GMs + $N_{\mathrm{NPC}}$ NPCs (NBA)
& $\checkmark$ & $\checkmark$ & $\checkmark$ & $\checkmark$ \\
\bottomrule
\end{tabularx}
\endgroup

\captionof{table}{
Interaction scales and evaluation designs of representative benchmarks.
}
\label{tab:benchmarks}

\vspace{3pt}
\begin{minipage}{\linewidth}
\footnotesize
\textit{Notes.}
$\checkmark$: explicit support;
$\triangle$: related comparisons of memory, frameworks, or incentives;
$\times$: no documented support under the stated definitions.
Counts include management agents and negotiation NPCs, excluding
evaluators and simulated consumer demand.
E-Commerce Bench provides a pool of 576 supplier NPCs with rule-based
economic decisions and LLM-rendered dialogue; each episode may contact
only a subset.
$N_{\mathrm{NPC}}$ counts player and agent NPCs in the NBA instance
of GM Bench.
\end{minipage}

\end{minipage}
\end{table}

\section{Model List}
\label{app:model-identifiers}

The model list comprises 31 distinct models, with 29 shared across all four sports. Basketball and baseball each use these 29 models plus one scripted baseline, yielding 30 participants to match the number of teams in the National Basketball Association (NBA) and Major League Baseball (MLB), respectively. American football and ice hockey additionally include GPT-5.4 Mini and GPT-5.4 Nano, yielding 31 models plus the scripted baseline, or 32 participants, to match the number of teams in the National Football League (NFL) and National Hockey League (NHL), respectively. The details are shown in Table~\ref{tab:model-identifiers}.

\begin{table}[ht]
\centering

\small
\renewcommand{\arraystretch}{1.2}

\renewcommand{\tabularxcolumn}[1]{m{#1}}

\begin{tabularx}{\linewidth}{
  @{}
  >{\raggedright\arraybackslash\bfseries}m{0.20\linewidth}
  >{\raggedright\arraybackslash}X
  @{}
}
\toprule
Group & \textbf{Model identifiers} \\
\midrule

OpenAI &
  \mbox{GPT-6 Astra};
  \mbox{GPT-6.1 Sol};
  \mbox{GPT-6 Luna};
  \mbox{GPT-5.6 Terra};
  \mbox{GPT-4o};
  \mbox{GPT-5.4 Mini};
  \mbox{GPT-5.4 Nano}. \\

Anthropic &
  \mbox{Claude Opus 5.5};
  \mbox{Claude Sonnet 5.5};
  \mbox{Claude Fable 5.1};
  \mbox{Claude Haiku 4.5}. \\

Gemini &
  \mbox{Gemini 3.1 Pro Preview};
  \mbox{Gemini 3.8 Flash};
  \mbox{Gemini 3.5 Flash-Lite}. \\

DeepSeek &
  \mbox{DeepSeek-V4-Pro};
  \mbox{DeepSeek-V4.1-Flash};
  \mbox{DeepSeek-R1}. \\

Qwen &
  \mbox{Qwen3.8-Max};
  \mbox{Qwen3.8-Flash};
  \mbox{Qwen3.8-27B}. \\

Grok &
  \mbox{Grok 4.7};
  \mbox{Grok 4.6}. \\

GLM &
  \mbox{GLM-5.3};
  \mbox{GLM-5.3-Flash}. \\

MiMo &
  \mbox{MiMo-V2.6-Pro};
  \mbox{MiMo-V2.6-Flash}. \\

Doubao &
  \mbox{Doubao-Seed-2.1-Pro};
  \mbox{Doubao-Seed-2.1-Turbo}. \\

Kimi    & \mbox{Kimi K3}. \\
MiniMax & \mbox{MiniMax M3}. \\
HY4     & \mbox{Hy4 preview}. \\

\midrule
Baseline & \mbox{Native GM strategy}. \\
\bottomrule
\end{tabularx}

\caption{Models used across the four sports.}
\label{tab:model-identifiers}

\end{table}

\section{Reproducibility Details}
\label{app:reproducibility}

\subsection{Evaluation Scope and Language}
The benchmark covers basketball, American football, ice hockey, and baseball. Section~5.1 describes five roster types for each sport: Balanced, Star, Twins, Offense, and Defense. All model interactions use English text. Table~1 reports 30 NBA entries, comprising 29 LLM deployments and one scripted baseline. Table~3 reports 32 entries for American football, 32 for ice hockey, and 30 for baseball, each including the scripted baseline. These counts describe the reported model-policy entries. Random-seed repetitions are reported separately below.

\subsection{Inference and Experimental Selection}
All evaluated models are accessed through their official APIs. Configurable reasoning effort is set to \texttt{high} for every model that supports this setting. The total API expenditure for the experiments is approximately USD~15,000, as reported in Section~6.1. The local Python interpreter is version 3.9.6.

\subsection{Runs and Result Aggregation}
Each reported full-horizon model evaluation consists of one complete five-season run covering 2022--2026. Token expenditure constrained the number of full-horizon repetitions. Multiple short-horizon tests with different random seeds were used for cross-checking. The reported full-horizon results remain single-run measurements; short-horizon tests are not counted as independent five-season replications.

Section~5.2 defines the aggregation used in the result tables. Elo is the arithmetic mean of season-end ratings over the five seasons, with ratings initialized to 1500 at the start of each season and updated using $K=20$ and a 400-point logistic scale. Terminal cash is measured at the end of the evaluation horizon. Table~2 reports regular-season wins by year and the difference in mean win rate between 2022-2023 and 2025-2026 ($w=2$), expressed in percentage points. The five seasons form a temporally dependent trajectory. The tables do not report confidence intervals across independent full-horizon runs.

\end{document}